\pdfoutput=1

\documentclass[11pt]{article}

\usepackage[]{EACL2023}

\usepackage{times}
\usepackage{latexsym}

\usepackage[T1]{fontenc}

\usepackage[utf8]{inputenc}

\usepackage{microtype}

\usepackage{xspace}
\usepackage{graphicx}
\usepackage{enumitem}
\usepackage{xcolor,colortbl}
\usepackage{tikz-dependency}
\usetikzlibrary{shapes,arrows}
\usepackage{amssymb}
\usepackage{float}
\usepackage{todonotes}

\usepackage{hyperref}       %
\usepackage{url}            %
\usepackage{booktabs}       %
\usepackage{amsfonts}       %
\usepackage{nicefrac}       %
\usepackage{natbib}
\usepackage{doi}

\usepackage{caption}
\usepackage{subcaption}

\usepackage{lscape}
\usepackage{soul} %
\usepackage{longtable} %
\usepackage{multirow}
\usepackage{dsfont}

\setlist[description]{leftmargin=\parindent,labelindent=0pt}

\newcommand{\blue}[1]{\textcolor{black}{#1}}
\newcommand{\camerareadyblue}[1]{\textcolor{black}{#1}}

\newcommand{\enquote}[1]{``#1''}

\newcommand{\sref}[1]{Sec.~\ref{#1}}
\newcommand{\aref}[1]{Appendix~\ref{#1}}

\newcommand{\tref}[1]{Table~\ref{#1}}
\newcommand{\fref}[1]{Figure~\ref{#1}}

\definecolor{alex}{rgb}{0.635,0.998,0.722}
\definecolor{anne}{rgb}{0.8,0.8,1}
\definecolor{sophie}{rgb}{0.998,0.722,0.635}
\definecolor{bill}{rgb}{0.4, 0.8, 0.4}
\definecolor{final}{rgb}{1, 1, 0.6}

\newcounter{example}

\usepackage{array}
\newcolumntype{L}[1]{>{\raggedright\let\newline\\\arraybackslash\hspace{0pt}}m{#1}}
\newcolumntype{C}[1]{>{\centering\let\newline\\\arraybackslash\hspace{0pt}}m{#1}}
\newcolumntype{R}[1]{>{\raggedleft\let\newline\\\arraybackslash\hspace{0pt}}m{#1}}
\usepackage{rotating}

\title{A Survey of Methods for Addressing Class Imbalance\\ in Deep-Learning Based Natural Language Processing}

\author{Sophie Henning$^{1,2}$ \hspace*{0.2cm} William Beluch$^1$ \hspace*{0.2cm} Alexander Fraser$^2$ \hspace*{0.2cm} Annemarie Friedrich$^1$ \\
  $^1$ Bosch Center for Artificial Intelligence, Renningen, Germany \\
  $^2$ Center for Information and Language Processing, LMU Munich, Germany \\
  \texttt{sophieelisabeth.henning|william.beluch@de.bosch.com} \\
  \texttt{fraser@cis.lmu.de} \\
  \texttt{annemarie.friedrich@de.bosch.com} \\}

\begin{document}
\maketitle
\begin{abstract}
    Many natural language processing (NLP) tasks are naturally imbalanced, as some target categories occur much more frequently than others in the real world.
    In such scenarios, current NLP models 
    tend to perform poorly on less frequent classes.
    Addressing class imbalance in NLP is an active research topic, yet, finding a good approach for a particular task and imbalance scenario is difficult.

    In this survey, the first overview on class imbalance in deep-learning based NLP, we first discuss various types of controlled and real-world class imbalance.
    Our survey then covers approaches that have been explicitly proposed for class-imbalanced NLP tasks or, originating in the computer vision community, have been evaluated on them.
    We organize the methods by whether they are based on sampling, data augmentation, choice of loss function, staged learning, or model design.
    Finally, we discuss open problems and how to move forward. %

\end{abstract}

\section{Introduction}
\label{sec:intro}

\begin{figure}[t]
    \begin{subfigure}[htb]{0.237\textwidth}
        \includegraphics[width=\textwidth]{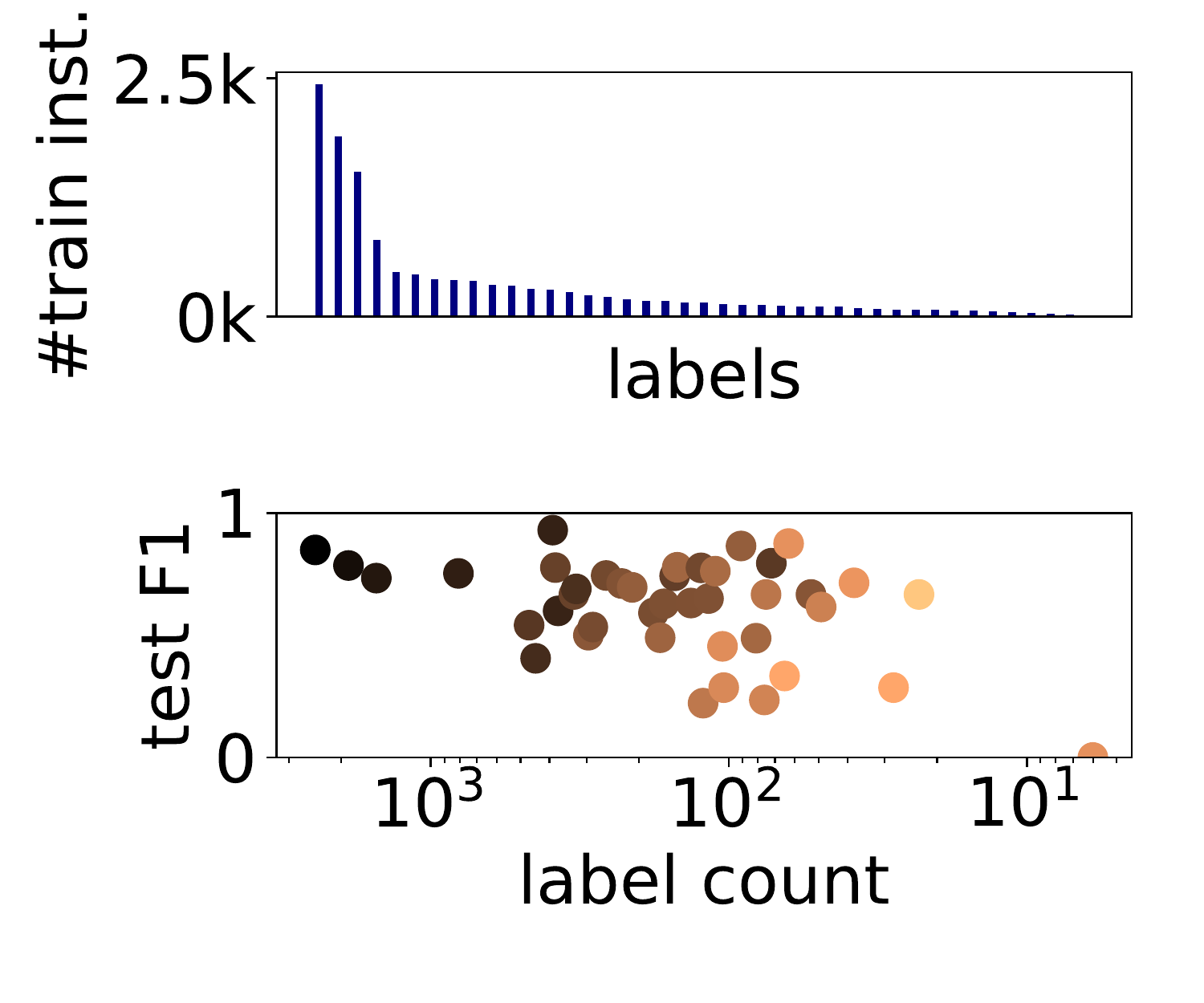}
        \vspace*{-8mm}
        \caption{Single-label \textbf{relation}\\ \hspace*{4mm}\textbf{classification} on TACRED\\ \hspace*{4mm}\citep{zhou2021improved}}
        \label{fig:single_label_text}
    \end{subfigure}
    \begin{subfigure}[htb]{0.237\textwidth}
        \includegraphics[width=\textwidth]{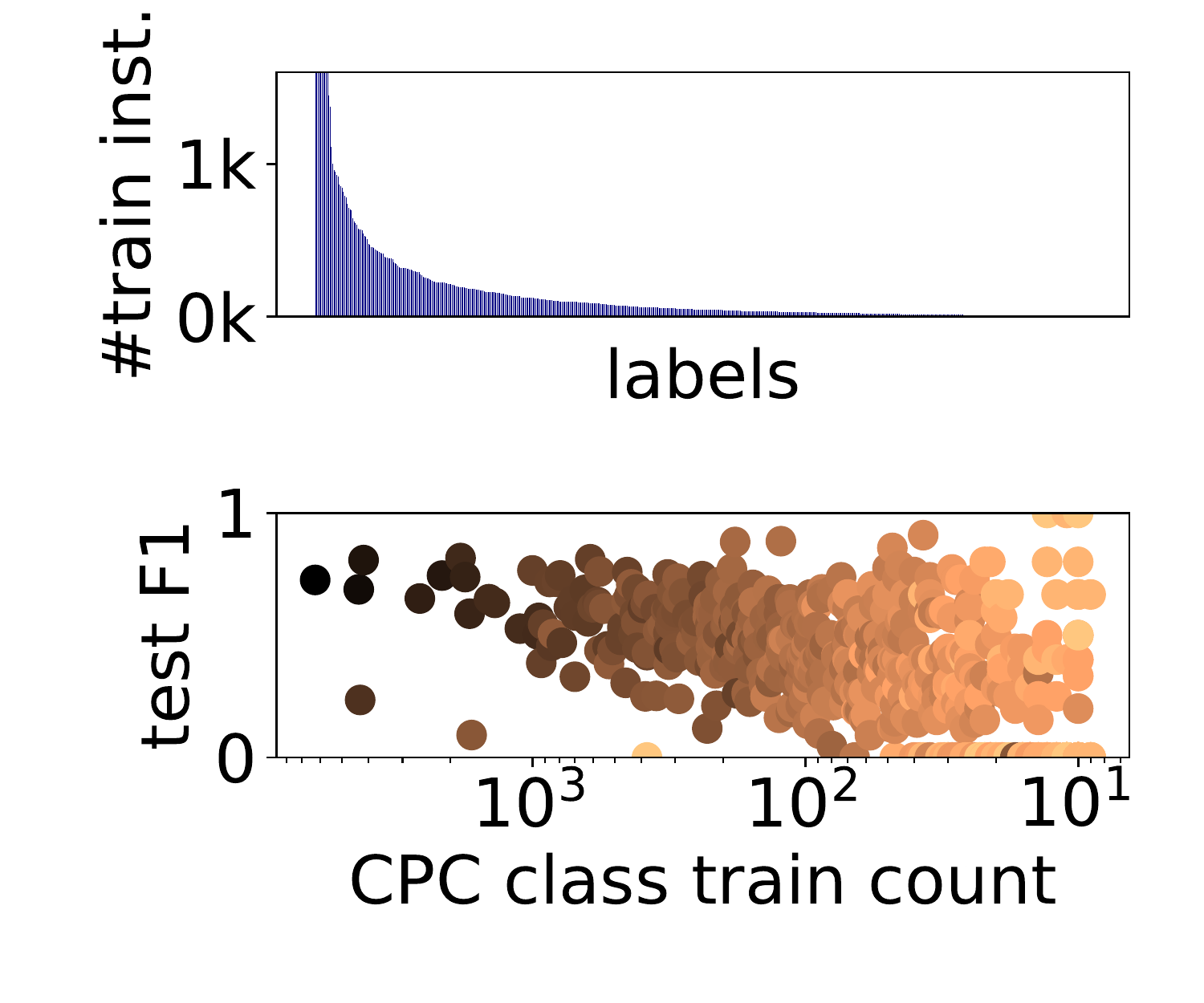}
        \vspace*{-8mm}
        \caption{Hierarchical multi-label\\\hspace*{4mm} \textbf{patent classification}\\ \hspace*{4mm} \citep{pujari2021multi}}
        \label{fig:multi_label_text}
    \end{subfigure}
    \begin{subfigure}[htb]{0.237\textwidth}
        \includegraphics[width=\textwidth]{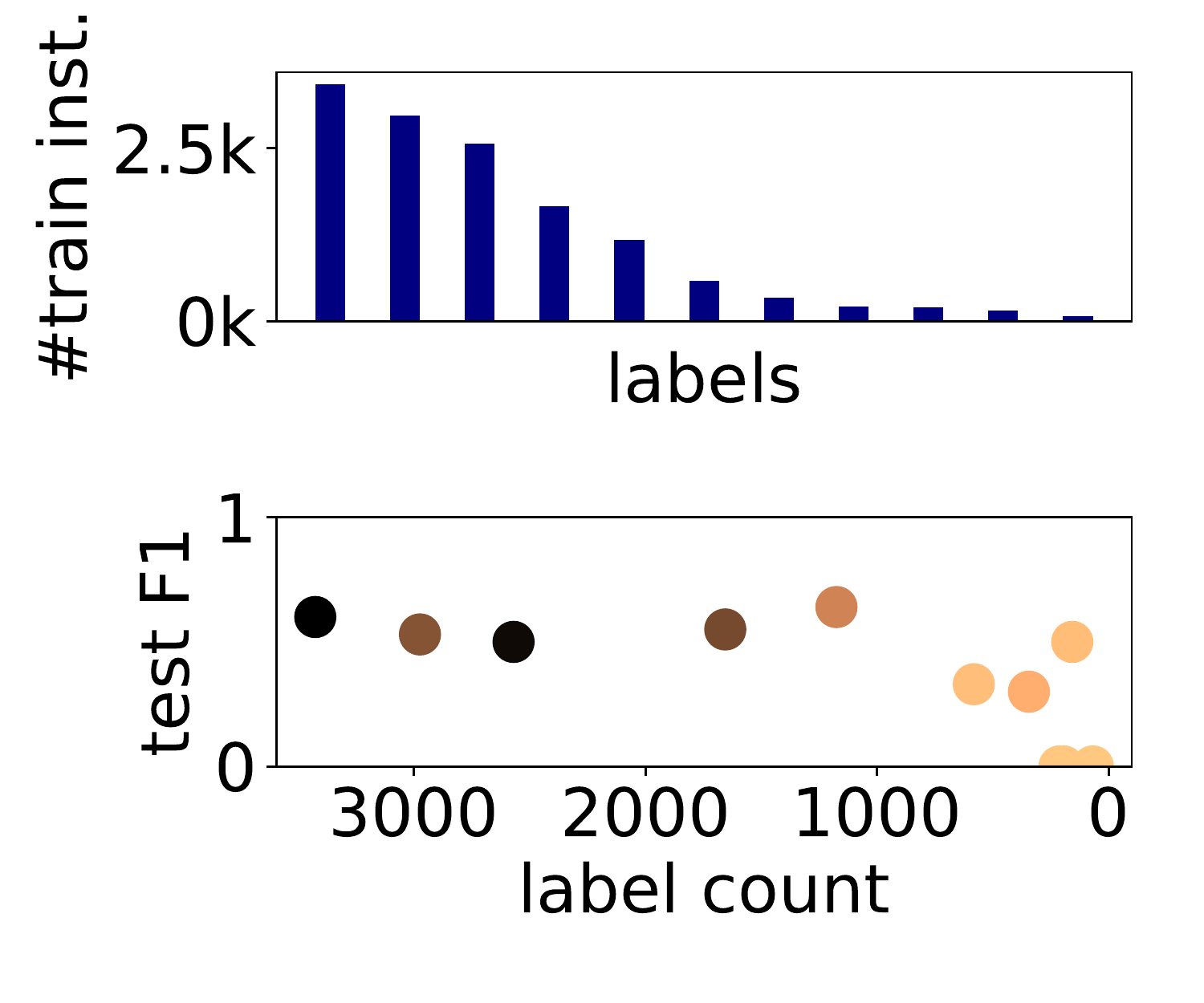}
        \vspace*{-8mm}
       \caption{Implicit \textbf{discourse rela-}\\\hspace*{4mm} \textbf{tion} classification (PDTB)\\\hspace*{4mm} \citep{shi-demberg-2019-next}}
        \label{fig:single_label_discourse}
    \end{subfigure}
    \begin{subfigure}[htb]{0.237\textwidth}
        \includegraphics[width=\textwidth]{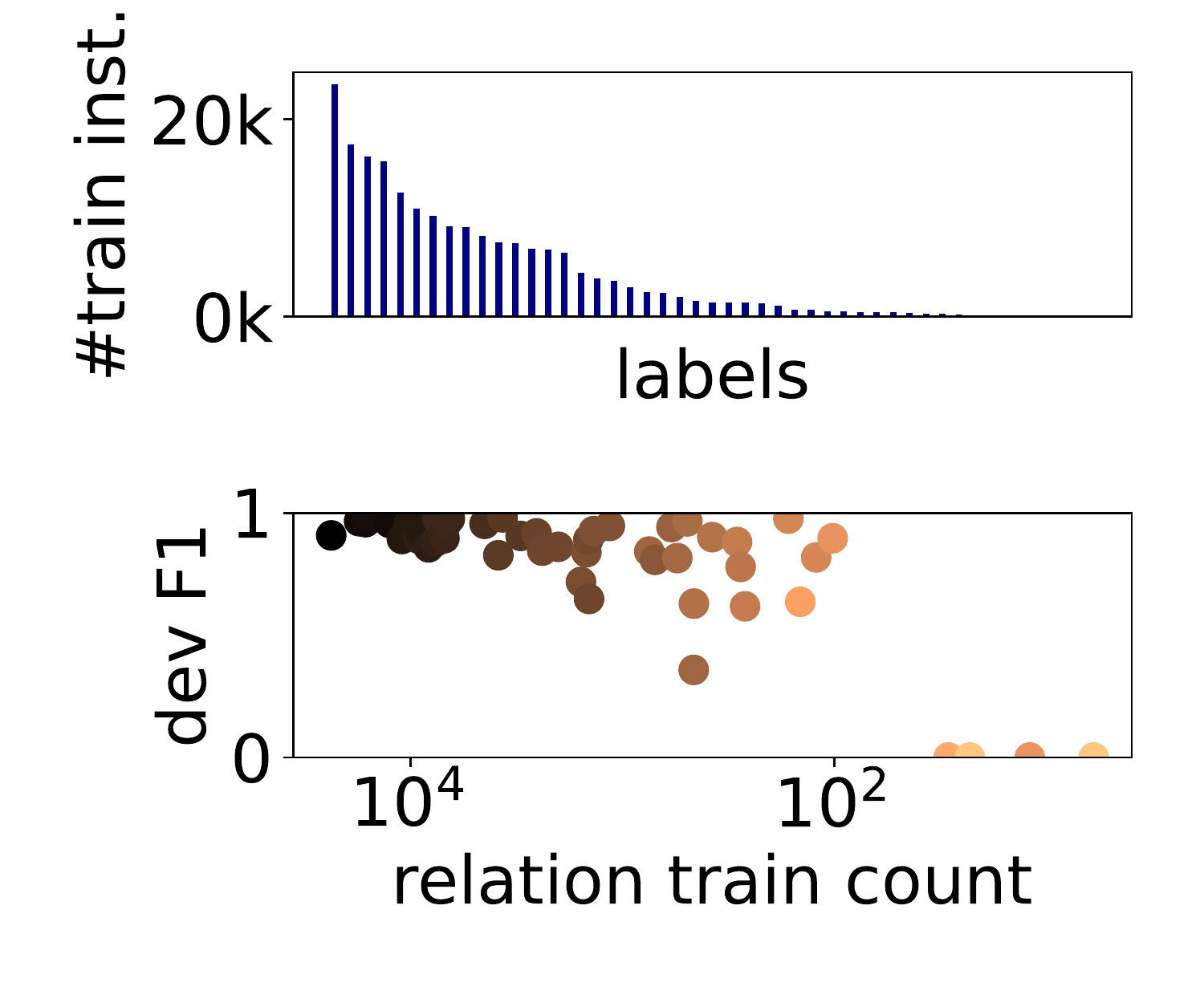}
        \vspace*{-8mm}
        \caption{UD \textbf{dependency parsing}\\\hspace*{4mm} using RoBERTa on EWT\\\hspace*{4mm} \citep{grunewald-etal-2021-applying}}
        \label{fig:dependency_parsing}
    \end{subfigure}
    \caption{\textbf{Class imbalance} has a negative effect on \textbf{performance} especially for minority classes in a variety of NLP tasks. Upper charts show label count distributions, lower part show test/dev F1 by training instance count (lighter colors indicate fewer test/dev instances). All models are based on transformers.} %
    \label{fig:teaser}
\end{figure}

Class imbalance is a major problem in natural language processing (NLP), because target category distributions are almost always skewed in NLP tasks.
As illustrated by \fref{fig:teaser}, this often leads to poor performance on minority classes.
Which categories matter is highly task-specific and may even depend on the intended downstream use.
Developing methods that improve model performance in imbalanced data settings has been an active area for decades \citep[e.g.,][]{bruzzone1997imbalanced, japkowicz2000learning, estabrooks-japkowicz-2001-mixture, park2002boosted, tan2005weightedKNN}, and is recently gaining momentum in the context of maturing neural approaches \citep[e.g.,][]{buda2018systematic, kang2020decoupling, li-etal-2020-dice, yang-etal-2020-hscnn,jiang2021textcut, spangher-etal-2021-multitask}.
The problem is exacerbated when classes overlap in the feature space \citep{lin-etal-2019-cost,tian2020graph}.
For example, in patent classification, technical categories differ largely in frequency, and the concepts mentioned in the different categories can be very similar.

\begin{figure*}[ht]
    \centering
    \begin{subfigure}[htb]{0.32\textwidth}
        \centering
        \includegraphics[height=25mm]{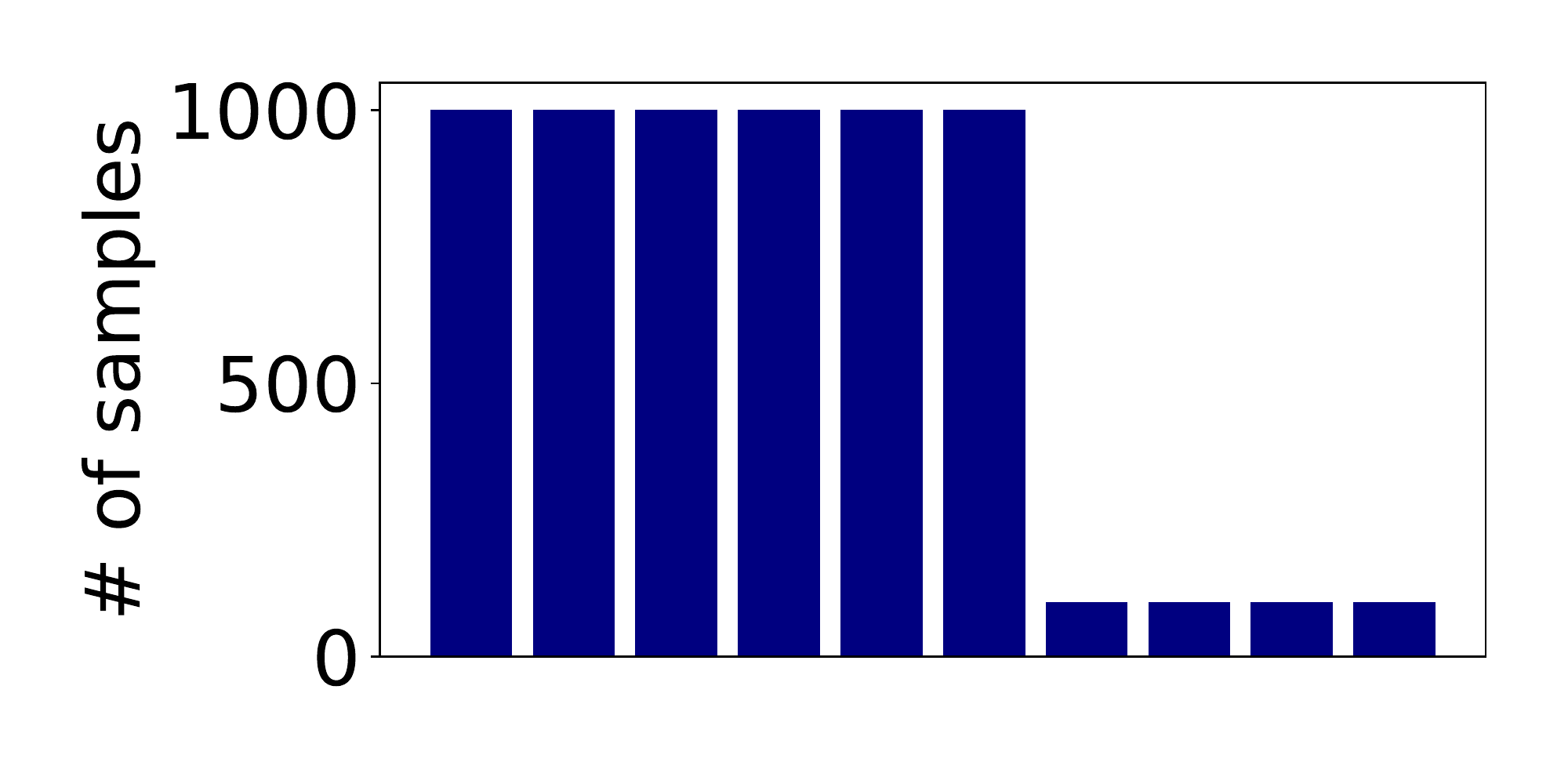}
        \caption{Step imbalance, $\mu=0.4$, $\rho=10$}
        \label{fig:step_imbalance}
    \end{subfigure}
    \begin{subfigure}[htb]{0.29\textwidth}
        \centering
        \includegraphics[height=25mm]{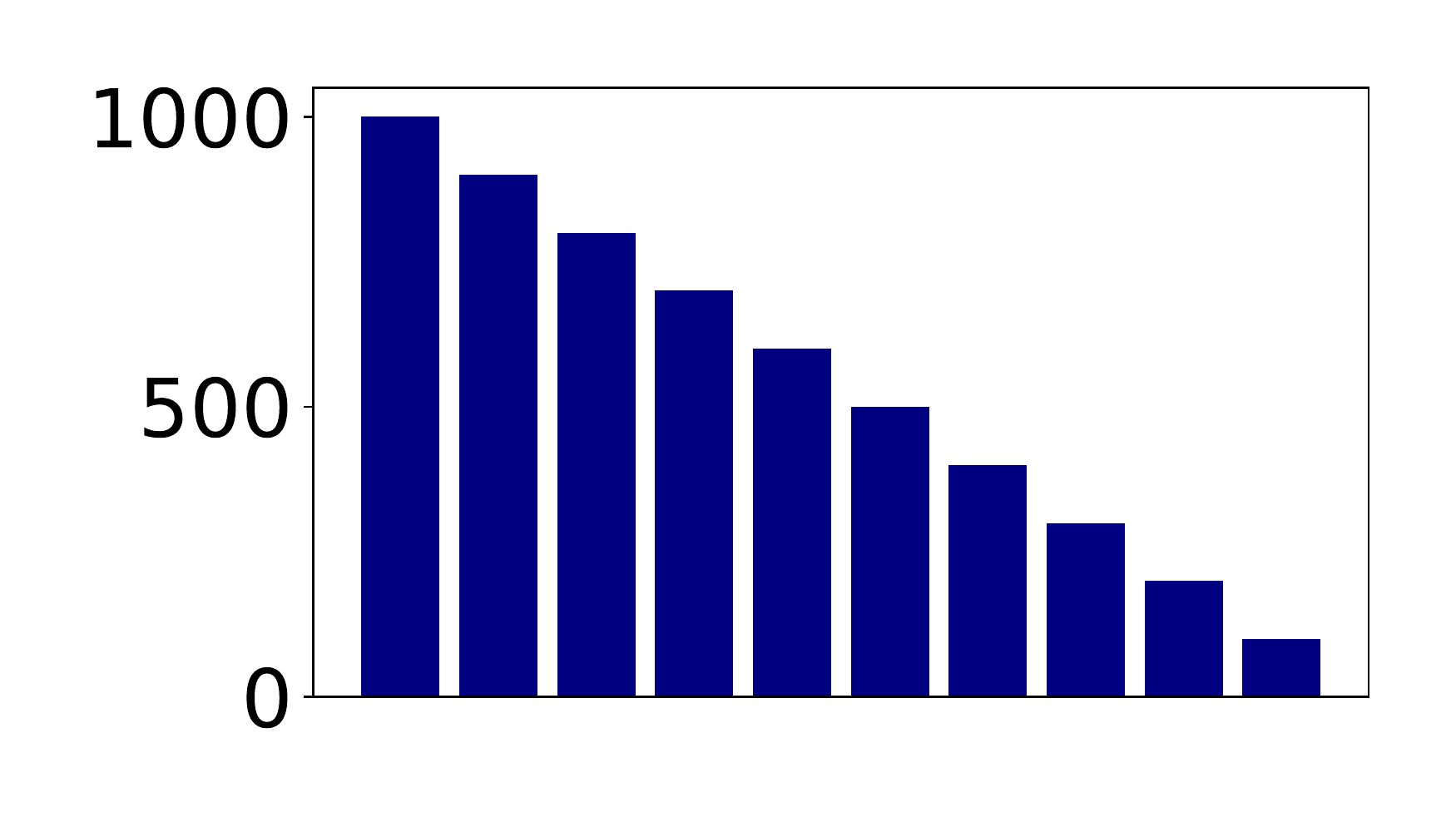}
        \caption{Linear imbalance, $\rho=10$}
        \label{fig:linear_imbalance}
    \end{subfigure}
    \begin{subfigure}[htb]{0.29\textwidth}
        \centering
        \includegraphics[height=25mm]{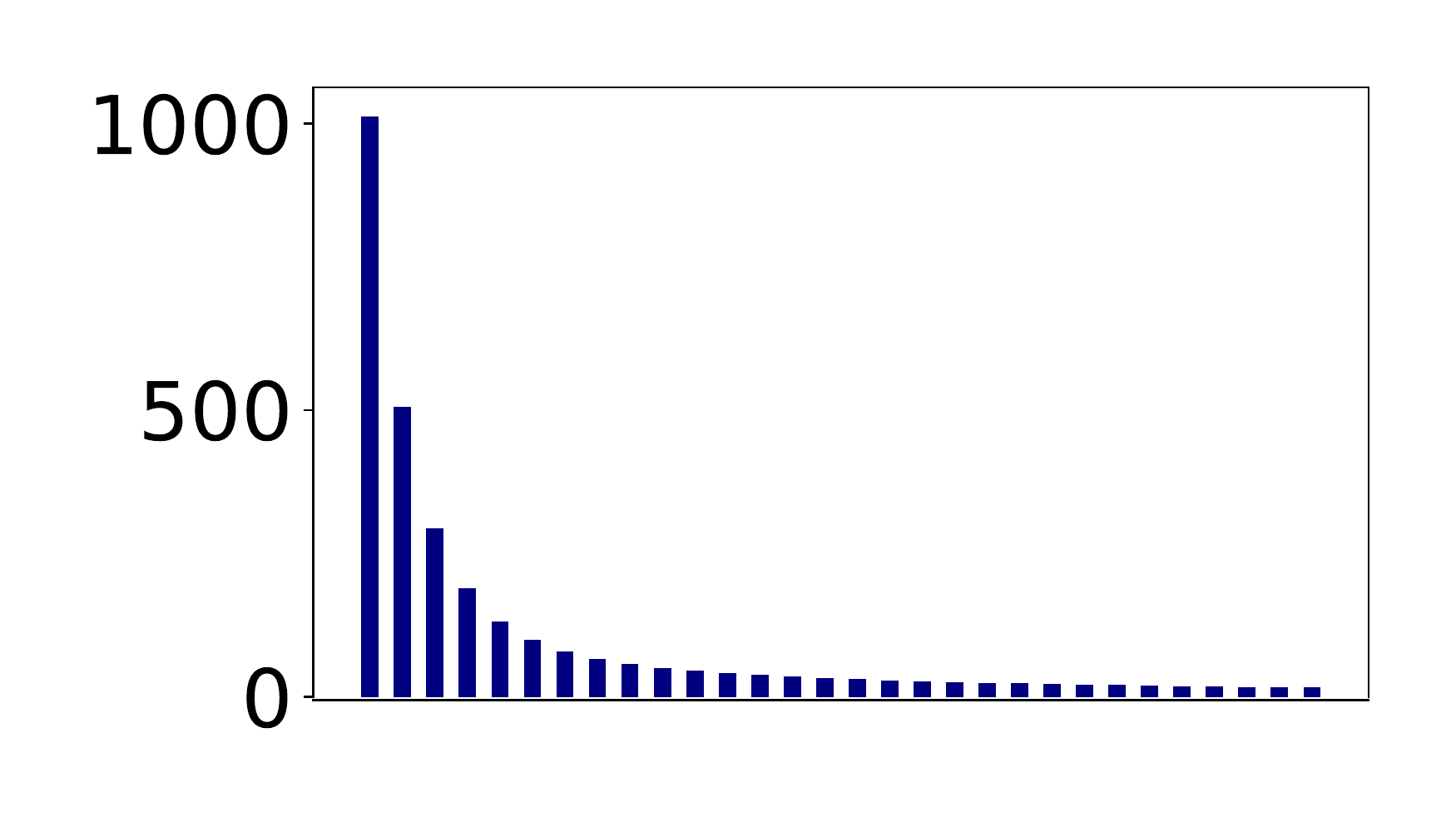}
        \caption{Long-tailed distribution}
        \label{fig:long_tailed}
    \end{subfigure}
    \caption{Instance counts per label follow different distributions: examples of \textbf{class imbalance types}.}
    \label{fig:example_distributions}
\end{figure*}

On a large variety of NLP tasks, transformer models such as BERT \citep{vaswani2017attention,devlin-etal-2019-bert} outperform both their neural predecessors and traditional models \citep{liu2019roberta,xie2020unsupervised,mathew2021hatexplain}.
Performance for minority classes is also often higher when using self-supervised pre-trained models \citep[e.g.,][]{li-scarton-2020-revisiting, niklaus-etal-2021-swiss}, which parallels findings from computer vision \citep{liu2022selfsupervised}.
However, the advent of BERT has not solved the class imbalance problem in NLP, as illustrated by \fref{fig:teaser}.
\citet{tanzer-etal-2022-memorisation} find \camerareadyblue{that on synthetically imbalanced named entity datasets with majority classes having thousands of examples, at least 25 instances are required to predict a class at all, and 100 examples to learn to predict it with some accuracy.}

Despite the relevance of class imbalance to NLP, related surveys only exist in the computer vision domain \citep{johnson_survey_2019,zhang2021deep}.
\blue{Incorporating methods addressing class imbalance can lead to performance gains of up to 20\%.
Yet, NLP research often overlooks how important this is in practical applications, where minority classes may be of special interest.}

Our contribution is to draw a clear landscape of approaches applicable to deep-learning (DL) based NLP. %
We set out with a problem definition (\sref{sec:definition}), and then organize approaches by whether they are based on sampling, data augmentation, choice of loss function, staged learning, or model design (\sref{sec:methods}).
\blue{
Our extensive survey finds that re-sampling, data augmentation, and changing the loss function can be relatively simple ways to increase performance in class-imbalanced settings and are thus straightforward choices for NLP practitioners.\footnote{\camerareadyblue{We provide practical advice on identifying potentially applicable class imbalance methods in the Appendix (\fref{fig:flowchart}).}}
While promising research directions, staged learning or model modifications often are implementation-wise and/or computationally  costlier.} %
Moreover, we discuss particular challenges of non-standard classification settings, e.g., imbalanced multi-label classification and catch-all classes, and provide useful connections to related computer vision work.
Finally, we outline promising directions for future research (\sref{sec:future}).

\textbf{Scope of this survey.}
We focus on approaches evaluated on or developed for neural methods.
Work from \enquote{traditional} NLP \citep[e.g.,][]{tomanek2009imbalance_crf, li2011imbalanced_maximum_entropy,li-nenkova-2014-addressing,kunchukuttan-bhattacharyya-2015-addressing} %
as well as %
Natural Language Generation \citep[e.g.,][]{nishino-etal-2020-reinforcement} and Automatic Speech Recognition \citep[e.g.,][]{winata2020multilingual_speech, deng2022asr} are not addressed in this survey. %
Other types of imbalances such as differently sized data sets of subtasks in continual learning \citep{ahrens2021drill} or imbalanced regression \citep{DBLP:conf/icml/YangZCWK21} are also beyond the scope of this survey.
In \sref{sec:model-design}, we briefly touch upon the related area of few-shot learning \citep{wang2020generalizing}.

\textbf{Related surveys.}
We review imbalance-specific data augmentation approaches in \sref{sec:data-augmentation}.
\citet{feng-etal-2021-survey} give a broader overview of data augmentation in NLP, \citet{hedderich-etal-2021-survey} provide an overview of low-resource NLP, and \citet{ramponi-plank-2020-neural} discuss neural domain adaptation.

\section{Problem Definition}
\label{sec:definition}

\textbf{Class imbalance} refers to a classification setting in which one or multiple classes (\textbf{minority classes}) are considerably less frequent than others (\textbf{majority classes}).
\camerareadyblue{More concrete definitions, e.g., regarding the relative share up to which a class is seen as a minority class, depend on the task, dataset and labelset size.}
Much research focuses on improving all minority classes equally while maintaining or at least monitoring majority class performance (e.g., \citealp{huang-etal-2021-balancing, yang-etal-2020-hscnn, spangher-etal-2021-multitask}).
We next discuss prototypical types of imbalance (\sref{sec:imbalance-types}) and then compare controlled and real-world settings (\sref{sec:settings}).

\subsection{Types of Imbalance}
\label{sec:imbalance-types}
To systematically investigate the effect of imbalance, \citet{buda2018systematic} define two \camerareadyblue{prototypical} types of label distributions, which we explain next.

\textbf{Step imbalance} is characterized by the fraction of minority classes, $\mu$, and the size ratio between majority and minority classes, $\rho$. 
Larger $\rho$ values indicate more imbalanced data sets.
In prototypical step imbalance, if there are multiple minority classes, all of them are equally sized; if there are several majority classes, they also have equal size.
\fref{fig:step_imbalance} shows a step-imbalanced distribution with $40\%$ of the classes being minority classes and an imbalance ratio of $\rho=10$.
\camerareadyblue{NLP datasets with a large catch-all class as they often arise in sequence tagging (see \sref{sec:settings}) or 
in
relevance judgments in retrieval models frequently resemble step-imbalanced distributions.}
The $\rho$ ratio has also been reported in NLP, e.g., by \citet{li-etal-2020-dice}, although more task-specific imbalance measures have been proposed, e.g., for single-label text classification \citep{tian2020graph}. 
\camerareadyblue{In \textbf{linear imbalance}, class size grows linearly with imbalance ratio $\rho$ (see \fref{fig:linear_imbalance}), as, e.g., in the naturally imbalanced SICK dataset for natural language inference \citep{marelli-etal-2014-sick}.}

\textbf{Long-tailed} label distributions (\fref{fig:long_tailed})
\camerareadyblue{are conceptually similar to linear imbalance.} 
\camerareadyblue{They} contain many data points for a small number of classes (\textit{head classes}), but only very few for the rest of the classes (\textit{tail classes}).
These distributions are common in computer vision tasks like instance segmentation \citep[e.g.,][]{gupta2019lvis}, but also in multi-label text classification, for example with the goal of assigning clinical codes \citep{mullenbach-etal-2018-explainable}, patent categories \citep{pujari2021multi}, or news and research topics \citep{huang-etal-2021-balancing}.

\subsection{Controlled vs. Real-World Class Imbalance}
\label{sec:settings}

Most real-world label distributions in NLP tasks %
do not perfectly match the prototypical distributions proposed by \citet{buda2018systematic}.
Yet, awareness of these settings helps practitioners to select appropriate methods for their data set or problem by comparing distribution plots.
Using synthetically imbalanced data sets, researchers can control for more experimental factors and investigate several scenarios at once.
However, evaluating on naturally imbalanced data provides evidence of a method's real-world effectiveness.
Some recent studies combine both types of evaluation (e.g., \citealp{tian-etal-2021-embedding, subramanian-etal-2021-fairness, jang2021sequential}). %

Many NLP tasks require treating a large, often heterogenous \textbf{catch-all class} that contains all instances that are not of interest to the task, while the remaining (minority) classes are approximately same-sized.
Examples include the \enquote{Outside} label in %
IOB
sequence tagging, or tweets that mention products in contexts that are irrelevant to the annotated categories \citep{adel-etal-2017-ranking}.
Such real-world settings often roughly follow a step imbalance distribution, with the additional difficulty of the catch-all class.

\subsection{Evaluation}
As \textit{accuracy} and \textit{micro}-averages mostly reflect majority class performance, choosing a good evaluation setting and metric is non-trivial.
It is also highly task-dependent: in many NLP tasks, recognizing one or all minority classes well is at least equally important as majority class performance.
For instance, non-hateful tweets are much more frequent in Twitter
\citep{waseem-hovy-2016-hateful}, but recognizing hateful content is the key motivation of hate speech detection.
Which classes matter may even depend on downstream considerations, i.e., the same named entity tagger might be used in one application where a majority class matters, and another where minority classes matter more. %
Several evaluation metrics exist that have been designed to account for class-imbalanced settings, but no de facto standard exists.
For example, \textit{balanced accuracy} \citep{brodersen2010balancedAcc} corresponds to the average of per-class recall scores.
It is often useful to record performance on \textit{all} classes and to report \textit{macro}-averages, which treat all classes equally.

\section{Methods for Addressing Class Imbalance in NLP}
\label{sec:methods}
In this section, we survey methods that either have been explicitly proposed to address class-imbalance issues in NLP or that have been empirically shown to be applicable for NLP problems. 
We provide an overview of which methods are applicable to a selection of NLP tasks in \aref{sec:appendix}.

\subsection{Re-Sampling}
\label{sec:sampling}
To increase the importance of minority instances in training, the label distribution can be changed by various sampling strategies.
Sampling can either be executed once or repeatedly during training \citep{dynamicSampling}.
In \textit{random oversampling} (\textbf{ROS}), a random choice of minority instances are duplicated, whereas in \textit{random undersampling} (\textbf{RUS}), a random choice of majority instances are removed from the dataset.
ROS can lead to overfitting
\blue{and increases training times.} %
RUS, however, discards potentially valuable data\blue{, but has been shown to work well in language-modeling objectives \citep{mikolov2013word_embeddings}.}

When applied in DL, ROS outperforms RUS both in synthetic step and linear imbalance \citep{buda2018systematic} and in binary and multi-class English and Korean text classification \citep{juuti-etal-2020-little, akhbardeh-etal-2021-handling, jang2021sequential}.
More flexible variants, e.g., re-sampling only a tunable share of classes  \citep{tepper-etal-2020-balancing} \blue{or interpolating between the (imbalanced) data distribution and an almost perfectly balanced distribution \citep{temperature_based_oversampling}}, can also further improve results.
\textit{Class-aware sampling} \citep[\textbf{CAS},][]{li2016class-aware}, also referred to as \textit{class-balanced sampling}, first chooses a class, and then an instance from this class.
Performance-based re-sampling during training, following the idea of \citet{dynamicSampling}, works well in multi-class text classification \citep{akhbardeh-etal-2021-handling}.

\textbf{Issues in multi-label classification.}
In multi-label classification, label dependencies between majority and minority classes complicate sampling approaches, as over-sampling an instance with a minority label may simultaneously amplify the majority class count \citep{charte2015mlros,huang-etal-2021-balancing}.
CAS also suffers from this issue, and additionally introduces within-class imbalance, as instances of one class are selected with different probabilities depending on the co-assigned labels \citep{wu2020distribution}.
Effective sampling in such settings is still an open issue.
Existing approaches monitor the class distributions during sampling \citep{charte2015mlros} or assign instance-based sampling probabilities \citep{gupta2019rfs, wu2020distribution}.

\subsection{Data Augmentation}
\label{sec:data-augmentation}
Increasing the amount of minority class data during corpus construction, e.g., by writing additional examples or selecting examples to be labeled using Active Learning, can mitigate the class imbalance problem to some extent \citep{cho-etal-2020-machines, ein-dor-etal-2020-active}.
However, this is particularly laborious in naturally imbalanced settings as it may require finding \enquote{the needle in the haystack,} %
or may lead to biased minority class examples, e.g., due to collection via keyword queries. 
Synthetically generating additional minority instances thus is a promising direction. %
In this section, we survey data augmentation methods that have been explicitly proposed to mitigate class imbalance and that have been evaluated in combination with DL. %

\textbf{Text augmentation}
generates new natural language instances of minority classes, ranging from simple string-based manipulations such as synonym replacements to Transformer-based generation.
\textit{Easy Data Augmentation} \citep[\textbf{EDA},][]{wei-zou-2019-eda}, which uses dictionary-based synonym replacements, random insertion, random swap, and random deletion,
has been shown to work well in class-imbalanced settings \citep{jiang2021textcut, jang2021sequential, juuti-etal-2020-little}.
\citet{juuti-etal-2020-little} generate new minority class instances for English binary text classification using EDA and embedding-based synonym replacements, and by adding a random majority class sentence to a minority class document.
They also prompt the pre-trained language model GPT-2 \citep{Radford2019LanguageMA} with a minority class instance to generate new minority class samples.
\citet{tepper-etal-2020-balancing} evaluate generation with GPT-2 on English multi-class text classification datasets, coupled with a flexible balancing policy (see \sref{sec:sampling}).

\blue{Similarly, \citet{gaspers20_intentclassification} combine machine-translation based text augmentation with dataset balancing to build a multi-task model.
Both the main and auxiliary tasks are German intent classification.
Only the training data for the latter is balanced and enriched with synthetic minority instances.}
\blue{In a long-tailed multi-label setting, \citet{zhang2022attention_text_augmentation} learn an attention-based text augmentation
that augments instances with text segments that are relevant to tail classes, leading to small improvements.} %
\blue{In general, transferring methods such as EDA or backtranslation to multi-label settings is difficult
\citep{zhang2022attention_text_augmentation, zhang2020data, tang2020multilabel}.}

\textbf{Hidden space augmentation} generates new instance vectors that are not directly associated with a particular natural language string, leveraging the representations of real examples. %
Using representation-based augmentations to tackle class imbalance is not tied to DL.
SMOTE \citep{chawla2002smote}, which interpolates minority instances with randomly chosen examples from their K-nearest neighbours, is popular in traditional machine learning \citep{fernandez2018smote}, but leads to mixed results in DL-based NLP \citep{ek-ghanimifard-2019-synthetic, tran-litman-2021-multi, wei2022reprint}.
Inspired by CutMix \citep{yun2019cutmix}, which cuts and pastes a single pixel region in an image, \textbf{TextCut} \citep{jiang2021textcut} randomly replaces small parts of the BERT representation of one instance with those of the other.
In binary and multi-class text classification experiments, TextCut improves over non-augmented BERT and EDA.

\textit{Good-enough example extrapolation} \citep[\textbf{GE3},][]{wei-2021-good} and \textbf{\textsc{Reprint}} \citep{wei2022reprint} also operate in the original representation space.
To synthesize a new minority instance, GE3 adds the vector representing the difference between a majority class instance and the centroid of the respective majority class to the mean of a minority class.
Evaluations on synthetically step-imbalanced English multi-class text classification datasets show improvements over oversampling and hidden space augmentation baselines. %
GE3 assumes that the distribution of data points of a class around its mean can be extrapolated to other classes, an assumption potentially hurting performance if the minority class distribution differs.
To account for this when subtracting out majority characteristics, \textsc{Reprint} performs a principal component analysis (PCA) for each class, leveraging the information on relevant dimensions during sample generation. %
This method usually outperforms GE3, with the cost of an additional hyperparameter (subspace dimensionality).

\textbf{MISO} \citep{tian-etal-2021-embedding} generates new instances by transforming the representations of minority class instances that are located nearby majority class instances.
They learn a mapping from minority instance vectors to \enquote{disentangled} representations, making use of mutual information estimators \citep{pmlr-v80-belghazi18a} to push these representations away from the majority class and closer to the minority class.
An adversarially-trained generator then generates minority instances using these disentangled representations. 
\citeauthor{tian-etal-2021-embedding} apply MISO in naturally and synthetically imbalanced English and Chinese binary and multi-class text classification with a single minority class.

\textbf{ECRT} \citep{chen2021supercharging} learns to map encoder representations (feature space) to a new space (source space) whose components are independent of each other given the class, assuming an invariant causal mechanism from source to feature space.
The independence enables them to generate new meaningful minority examples by permuting or sampling components in the source space, resulting in  %
\blue{medium} improvements on a large multi-label text classification dataset with many labels.

Further related work exists in the area of \textbf{transfer learning} \citep{ruder-etal-2019-transfer}, e.g., from additional datasets that provide complementary information on minority classes.
For instance, \citet{spangher-etal-2021-multitask} achieve small gains by manually selecting auxiliary datasets to improve imbalanced sentence-based discourse classification. %
However, complementary datasets have to be retrieved for each application, and task loss coefficients have to be tuned.
Adapting methods to predict useful transfer sources \citep{lange-etal-2021-share} might help alleviate these problems.

\subsection{Loss Functions}
\label{sec:loss}
Standard \textit{cross-entropy loss} (\textbf{CE}) is composed from the predictions for instances that carry the label in the gold standard, which is why the resulting classifiers fit the minority classes less well.
In this section, we summarize loss functions designed for imbalanced scenarios.
They either re-weight instances by class membership or prediction difficulty, or explicitly model class margins to change the decision boundary.
Throughout this section, we use the variables and terms as shown in \tref{tab:loss_functions}.

\paragraph{Losses for Single-Label Scenarios.}
\textit{Weighted cross-entropy} (\textbf{WCE}) uses class-specific weights \camerareadyblue{$\alpha_j$} that are tuned as hyperparameters or set to the inverse class frequency \citep[e.g.,][]{adel-etal-2017-ranking, tayyar-madabushi-etal-2019-cost, li-xiao-2020-syrapropa}.
While WCE treats all instances of one class in the same way, \textit{focal loss} \citep[\textbf{FL},][]{lin2017focal} down-weights instances for which the model is already confident \camerareadyblue{(implemented with the $\left(1-p_j\right)^{\beta}$ coefficient).}
FL can of course also be used with class weights.
Instead of mimicking accuracy like CE, \textit{dice loss} \citep[\textbf{Dice},][]{milletari2016vnet} tries to capture class-wise F1 score, with \camerareadyblue{the} predicted probability $p_j$ proxying precision and \camerareadyblue{the} ground truth indicator $y_j$ proxying recall.
\textit{Self-adjusting dice loss} \citep[\textbf{ADL},][]{li-etal-2020-dice} combines confidence-based down-weighting \camerareadyblue{via $1-p_j$} with Dice loss.
For sequence labeling, QA and matching on English and Chinese datasets, Dice performs better than FL and ADL.

Rather than re-weighting instances, \textit{label-distribution-aware margin loss} \citep[\textbf{LDAM},][]{cao2019ldam}, essentially a smooth hinge loss with label-dependent margins \camerareadyblue{$\Delta_j$}, aims to increase the distance of the minority class instances to the decision boundary with the aim of better generalization for these classes.
\citeauthor{cao2019ldam}'s evaluation largely focuses on computer vision, but they also report results 
for
LDAM on a synthetically imbalanced version of the IMDB review dataset \citep{maas-etal-2011-learning}, achieving a much lower error on the minority class %
than vanilla CE or CE with re-sampling or re-weighting.
\citet{subramanian-etal-2021-fairness} propose LDAM variants that consider bias related to socially salient groups (e.g., gender-based bias) in addition to class imbalance\blue{, evaluating them on binary text classification}. %

\begin{table*}[t]
    \centering
    \setlength\tabcolsep{3pt}
    \renewcommand{\arraystretch}{1.1}
    \footnotesize
    \begin{tabular}{r|rl|}
    \toprule
     \parbox[t]{2mm}{\multirow{3}{*}{\rotatebox[origin=c]{90}{\small \textbf{Single-label}\hspace*{5mm}}}} &    {\footnotesize \textbf{CE}} &  $-\sum_{j=1}^C y_j \log p_j$ \hspace*{17mm} {\footnotesize \textbf{WCE}} $-\sum_{j=1}^C \alpha_j y_j \log p_j$\\
        & {\footnotesize \textbf{FL}} & $-\sum_{j=1}^C y_j (1-p_j)^\beta \log p_j  $\\
        & {\footnotesize \textbf{Dice}} & $\sum_{j=1}^C 1-\displaystyle\frac{2p_j y_j + \gamma}{p_j^2+y_j^2+\gamma}$ \hspace*{9mm} {\footnotesize \textbf{ADL}} $\sum_{j=1}^C 1- \displaystyle\frac{2(1-p_j) p_j y_j + \gamma}{(1-p_j)p_j+y_j+\gamma} $ \\
         & {\footnotesize \textbf{LDAM}} & $-\sum_{j=1}^C y_j \log \displaystyle \frac{\exp(z_j-\Delta_j)}{\exp(z_j-\Delta_j)+\sum_{l\neq j} \exp(z_l)}$ with $\Delta_j = K / n_j^{1/4}$\\
        & {\footnotesize \textbf{RL}} & $\mathds{1}(gt \neq A) \log (1+\exp(\rho (m^+ - z_{gt}))) + \log(1+\exp(\rho (m^- - z_{c^-})))$\\
        \midrule
        \parbox[t]{2mm}{\multirow{3}{*}{\rotatebox[origin=c]{90}{\small \textbf{Multi-label}\hspace*{2mm}}}}& {\footnotesize \textbf{BCE}} & $-\sum_{j=1}^C [y_j \log p_j + (1-y_j) \log(1-p_j)]$ \\
        & {\footnotesize \textbf{WBCE}} & $-\sum_{j=1}^C \alpha_j [y_j \log p_j + (1-y_j) \log(1-p_j)]$ \\
        & {\footnotesize \textbf{FL}} & $-\sum_{j=1}^C [y_j (1-p_j)^\beta \log p_j + (1-y_j) p_j^\beta \log(1-p_j)]$ \\
        & {\footnotesize \textbf{DB}} & $-\sum_{j=1}^C [ y_j \hat{\alpha}_j (1-q_j)^\beta \log q_j + (1-y_j) \hat{\alpha}_j \frac{1}{\lambda} q_j^\beta \log (1-q_j) ]$\\
        & & with  $q_j=y_j\sigma(z_j-v_j)+(1-y_j)\sigma(\lambda(z_j-v_j))$\\
        \bottomrule
    \end{tabular}
       \renewcommand{\arraystretch}{1.07}
           \scriptsize
    \begin{tabular}{ll}
    \toprule
         $C$ & number of classes\\
         $y$ & target vector\\
          $p$ & model prediction vector\\
          $\alpha$ & class weights\\
          $\beta$ & tunable focusing parameter\\
          $z$ & model logits vector\\
          $\gamma$ & smoothing constant\\
          $n_j$ & size of class $j$ \\
          $K$ & label-independent constant\\
          $gt$ & index of ground-truth class\\
          $m^+$ & margin to correct class \\
          $m^-$ & ... to most competitive incorrect class\\
          $A$ & special catch-all class\\
          $c^-$ & index of largest non-$gt$ logit\\
         $\lambda$ & scaling factor\\
         $\hat{\alpha}_j$ & instance-specific class weights\\ %
          $v_j$ &  class-specific bias\\
          \bottomrule
    \end{tabular}
    
    \caption{Overview of \textbf{loss functions} formulated for one instance. See \aref{sec:appendix} for references/implementations. }
    \label{tab:loss_functions}
\end{table*}

In settings with a large artificial and potentially heterogeneous \textbf{catch-all class} (see \sref{sec:settings}), many areas of the space contain representations of the catch-all class.
Here, vanilla LDAM might be an appropriate loss function as it encourages larger margins for minority classes.
In such cases, \textit{ranking losses} (\textbf{RL}) can also be effective to incentivize the model to only pay attention to \enquote{real} classes. %
On an imbalanced English multi-class dataset with a large catch-all class, \citet{adel-etal-2017-ranking} find 
a ranking loss
introduced by \citet{dos-santos-etal-2015-classifying} 
improves
over CE and WCE.
For minority classes, this loss function maximizes the score of the correct label \camerareadyblue{$z_{gt}$} while at the same time minimizing the score of the highest-scoring incorrect label \camerareadyblue{$z_{c^-}$}. %
For the catch-all class \camerareadyblue{$A$}, only \camerareadyblue{$z_{c^-}$} %
is minimized; \camerareadyblue{$z_{gt}$} %
is ignored.
Similarly, \citet{hu2022event} apply class weights only to non-catch-all classes.

\paragraph{Losses for Multi-Label Scenarios.}
In multi-label classification, each label assignment %
can be viewed as
a binary decision, hence \textit{binary cross-entropy} (\textbf{BCE}) is often used here.
Under imbalance, two issues arise.
First, although class-specific weights have been used with BCE \citep[e.g.,][]{yang-etal-2020-hscnn}, their effect on minority classes is less clear than in the single-label case. 
For each instance, all classes contribute to BCE, %
\camerareadyblue{with} the labels \textit{not} assigned to the instance (called \textit{negative classes}) \camerareadyblue{included via $(1-y_j) \log(1-p_j)$}.
Thus, if \camerareadyblue{\textit{weighted binary cross-entropy} (\textbf{WBCE})} uses a high weight for a class, it also increases the importance of negative instances for a minority class, which may further encourage the model \camerareadyblue{to} \textit{not} predict this minority class.

To leverage class weights more effectively in BCE, one option is to only apply them to the loss of positive instances as proposed for multi-label image classification
\citep{kumar2018boosted}.
Related work includes uniformly upweighting positive instances of \textit{all} classes in hierarchical multi-label text classification \citep[e.g.,][]{rathnayaka2019gated}.
An approach to multi-label emotion classification by \citet{yilmaz2021multilabel_weighting} performs training time balancing by adapting FL such that for a given mini-batch the loss over all instances in this mini-batch has exactly the same value for every class.

If a classifier already correctly predicts a negative class for an instance, the loss can be further decreased by reducing the respective label's logits.
In CE, due to the softmax that uses the logits of all classes, the impact of this effect becomes minor once the logit for the correct class is much larger than those of the other classes.
However, the problem is more severe in BCE \citep{wu2020distribution}, as logits are treated independently.
As minority labels mostly occur as negative classes, this logit suppression leads to a bias in the decision boundary, making it less likely for minority classes to be predicted.
To tackle this issue and based on a multi-label version of FL, \citet{wu2020distribution} propose \textit{distribution-balanced loss} (\textbf{DB})  for object detection, adding \textit{Negative Tolerant Regularization} for the loss for negative classes by transforming the logits of positive and negative classes differently (see $q_j$ in \tref{tab:loss_functions}). %
This regularization imposes a sharp drop in the loss function for negative classes once the respective logit is below a threshold.
\blue{Moreover,} DB introduces instance-specific class weights %
$\hat{\alpha}$ %
to account for imbalances caused by class-aware sampling (see \sref{sec:sampling}) in multi-label scenarios.
These weights reflect the frequency of a class and the quantity and frequency of the positive labels of the instance.
\citet{huang-etal-2021-balancing} have shown large improvements of DB over BCE even when using uniform sampling on two long-tailed multi-label English text classification datasets.

\camerareadyblue{While \citet{cao2019ldam} propose and theoretically justify LDAM for single-label classification only, it has been successfully applied to multi-label text classification as well \citep{biswas2021ldam_multilabel}.} 
\citet{ferreira-vlachos-2019-incorporating} show that applying a cross-label dependency loss \citep{yeh2017learning,zhang2006} can be helpful for multi-label stance classification.
Similarly, \citet{lin-etal-2019-cost} introduce a label-confusion aware cost factor into their loss function.
The adaptive loss of \citet{suresh-ong-2021-negatives} integrates inter-label relationships into a contrastive loss \citep{khosla2020supervised}, which compares the score of a positive example with the distance to that of other positive and negative examples in order to push its representation closer to the correct class and further away from the wrong class(es).
The resulting loss function learns how to increase the weight of confusable negative labels relative to other negative labels.
\blue{Combining label-confusion aware loss functions with class weighting techniques is a promising research direction.}

\paragraph{Re-Sampling vs. Loss Functions.}
\camerareadyblue{
Re-sampling and loss functions that are specifically designed for class-imbalanced settings are based on the same idea of increasing the importance of minority instances.
Re-sampling is conceptually simpler and has a direct impact on training time, e.g., oversampling may cause a considerable increase.
By contrast, the loss functions explained above are more flexible, e.g., by modeling desirable properties of margins, but also mostly harder to interpret.}

\subsection{Staged Learning}
\label{sec:staged-learning}
One approach to finding a good trade-off between learning features that are representative of the underlying data distribution and reducing the classifier's bias towards the majority class(es) is to perform the training in several stages.
\textit{Two-staged training} is common in imbalanced or data-scarce computer vision tasks \citep[e.g.,][]{wang2020tfa, wang2020simcal, zhang2021distribution_alignment}.
The first stage usually performs standard training in order to train or fine-tune the feature extraction network.
Later stages may freeze the feature extractor and re-train the classifier layers using special methods to address class imbalance, e.g., using more balanced data distributions or specific losses.
For example, \citet{cao2019ldam} find their LDAM loss to be most effective when the training happens in two stages. %
In NLP, deep-learning models are usually based on pre-trained neural text encoders or word embeddings. %
Further domain-specific pre-training before starting the fine-tuning stage(s) can also be effective \citep{gururangan-etal-2020-dont}.

Several NLP approaches that fall under \textit{staged learning} are directly inspired by computer vision research.
In the context of long-tailed image classification, \citet{kang2020decoupling} find that class-balanced sampling (see \sref{sec:sampling}) helps when performing single-stage training, but that in their two-stage \textit{classifier re-training} (\textbf{cRT}) method, using the original distribution in the first stage is more effective than class-balanced sampling.
cRT employs the latter only in the second stage after freezing the representation weights.
\citet{yu2020devil} perform a similar decoupling analysis on long-tailed relation classification, essentially confirming \citet{kang2020decoupling}'s %
results on this NLP task with respect to the re-sampling strategies.
Additionally, they find that loss re-weighting under this analysis behaves similar to re-sampling, i.e., it leads to worse performance when applied during representation learning, but boosts performance when re-training the classifier.
\citet{hu2022event} successfully leverage \citeauthor{kang2020decoupling}'s ideas for event detection, where both trigger detection and trigger classification suffer from class imbalance.

\citet{jang2021sequential} model imbalanced classification as a continual learning task with $k$ stages where the data gradually becomes more balanced %
(\textit{sequential targeting}, \textbf{ST}).
The first stage contains the most imbalanced subset, and then the degree of imbalance decreases, with the last stage presenting the most balanced subset.
The training objective encourages both good performance on the current stage and keeping information learnt in previous stages.
Their experiments include binary and ternary English and Korean text classification.
Active Learning (AL), which contains several stages by definition, has also been shown to boost performance of BERT models for minority classes \citep{ein-dor-etal-2020-active}.
For a discussion about AL and deep learning, see \citet{schroeder2020al}. %

\subsection{Model Design}
\label{sec:model-design}
The methods described so far are largely independent of model architecture.
This section describes model modifications that aim at improving performance in imbalanced settings. %

Observing that the weight vectors for smaller classes have smaller norms in standard joint training compared to  staged-learning based cRT (see \sref{sec:staged-learning}), \citet{kang2020decoupling}
normalize the classifier weights directly in one-staged training using a hyperparameter $\tau$ to control the normalization \enquote{temperature} (\textbf{$\tau$-norm}).
$\tau$-norm achieves similar or better performance than cRT in long-tailed image classification and outperforms cRT in relation extraction, but cRT works better for named entity recognition and event detection \citep{nan-etal-2021-uncovering}.

\textbf{SetConv} \citep{gao-etal-2020-setconv} and \textbf{ProtoBERT} \citep{tanzer-etal-2022-memorisation} learn representatives for each class using \textit{support sets} and classify an input (the \textit{query}) based on its similarity to these representatives.
SetConv applies convolution kernels that capture intra- and inter-class correlations to extract class representatives.
ProtoBERT uses class centroids in a learned BERT-based feature space, treating the distance of any instance to the catch-all class as just another learnable parameter. %
At each training step, SetConv uses only one instance per class in the query set, but preserves the original class imbalance in the support set, whereas ProtoBERT uses fixed ratios.
In the respective experimental studies, ProtoBERT performs better than using a standard classification layer on top of BERT for minority classes in NER if less than 100 examples are seen by the model, while SetConv excels in binary text classification with higher degrees of imbalance and in multi-class text classification.

The \textbf{HSCNN} model \citep{yang-etal-2020-hscnn} uses class representatives only for the classification of tail classes, while head classes are assigned using a standard text CNN \citep{kim-2014-convolutional}.
HSCNN learns label-specific similarity functions, extracting instance representations from the pre-final layers of two copies of the original CNN, and assigns a tail class if the similarity to the class representative (computed as the mean of 5 random support instances) exceeds 0.5.
On tail classes, HSCNN consistently improves over the vanilla CNN.

In addition, there exist a number of \textbf{task-specific solutions}.
\citet{prange-etal-2021-supertagging} propose to construct CCG supertags from predicted tree structures rather than treating the problem as a standard classification task.
In order to recognize implicit positive interpretations in negated statements in a class-imbalanced dataset,
\citet{van-son-etal-2018-scoring} argue that leveraging information structure could be one way to improve inference. %
\textit{Structural causal models} (\textbf{SCM}s) have been applied to imbalanced NLP tasks, encoding task-specific causal graphs \citep[e.g.,][]{nan-etal-2021-uncovering}.
Similarly, \citet{wu-etal-2021-discovering} causally model how bias in long-tailed corpora affects topic modeling \citep{blei2003topic} and use this to %
improve training of a variational autoencoder.

A research area closely related to class imbalance is \textbf{few-shot learning} \citep[FSL,][]{wang2020generalizing}, which aims to learn classes based on only very few training examples.
Model ideas from FSL can be leveraged for long-tailed settings,
e.g., by making use of relational information about class labels in the form of knowledge graph embeddings or other forms of embedding hierarchical relationships between labels \citep{han-etal-2018-hierarchical,zhang-etal-2019-long}, or computing label-specific representations \citep{mullenbach-etal-2018-explainable}.

\section{Insights and Future Directions}
\label{sec:future}
We have provided a comprehensive, concise and structured overview of current approaches to dealing with class imbalance in DL-based NLP. %

\textbf{What works (best)?}
\camerareadyblue{As there is no established benchmark for class-imbalanced settings, evaluation results are hard to compare across papers.
In general, re-sampling or changing the loss function may lead to small to moderate gains.
For data augmentation approaches, the reported performance increases tend to be larger than for re-sampling or new loss functions.
The effects of staged training or modifications of the model vary drastically, ranging from detrimental to very large performance gains.
}

\camerareadyblue{Hence,}
\blue{
re-sampling, data augmentation, and changing the loss function are straightforward choices in class-imbalanced settings.
Approaches based on staged learning or model design may sometimes outperform them, but often come with 
a
higher implementation or computational cost.
\camerareadyblue{For a practical decision aid and potential application settings of some class imbalance methods}, see \camerareadyblue{\fref{fig:flowchart} in} \aref{sec:practical} \camerareadyblue{and \tref{tab:overview} in \aref{sec:appendix}}.}

\textbf{How should we report results?}
Much NLP \blue{research} only reports aggregate statistics \citep{harbecke-etal-2022-micro}, making it hard to judge the impact on improvements by class, \blue{which is often important in practice.}
We \blue{thus} argue that NLP researchers should \textit{always} report per-class statistics, \blue{e.g., as in \fref{fig:teaser}.
Open-sourcing spreadsheets with the exact numbers} would enable the community to \camerareadyblue{compare systems more flexibly from multiple angles, i.e., with respect to 
whichever
class(es) matter in a particular application scenario, and to }re-use \blue{this} data \blue{in} research on class imbalance.
Reviewers should 
also
value works that analyze performance for relevant minority classes rather than focusing largely only 
on 
overall
accuracy improvements.

A main hindrance to making progress on class imbalance in computer vision and NLP alike is that experimental results are often hard to compare \citep{Johnson2019DeepLA,johnson_effects_2020}.
A first important step would be 
to not restrict baselines to methods of 
the
same type, e.g., a new data augmentation approach should not only compare to other data augmentation methods, but also to using loss functions for class imbalance.
Establishing a shared and systematic benchmark of a diverse set of class-imbalanced NLP tasks would be highly beneficial for both researchers and practitioners.

\blue{\textbf{How can we move forward?}
Most work on class-imbalanced NLP has focused on single-label text classification. Finding good solutions for multi-label settings is still an open research challenge.}
\blue{Class imbalance also poses problems in NLP tasks such as sequence labeling or parsing, and we believe that the interaction of structured prediction models with methods to address class imbalance is a promising area for future research.}
\blue{
Moreover, %
we need to study
how class imbalance methods affect prediction calibration in order to provide reliable confidence estimates.}

\blue{
In general, inspiration for new model architectures could for example be drawn from approaches developed for few-shot learning \citep{wang2020generalizing}.
Recently, prompting \citep{Radford2019LanguageMA} has emerged as a new paradigm in NLP, which performs strongly in real-world few-shot settings \citep{schick-schutze-2022-true}.
Methods that improve worst-case performance under distribution shift \citep[e.g.,][]{Sagawa2020Distributionally} might also be applied to improve minority-class performance.} %

\section*{Acknowledgements}
We thank the anonymous reviewers for their valuable comments, and Heike Adel, Stefan Grünewald, Subhash Pujari, and Timo Schrader for providing data for our teaser image.
We thank them and Talita Anthonio, Mohamed Gad-Elrab, Lukas Lange, Stefan Ott, Robert Schmier, Hendrik Schuff, Daria Stepanova, Jannik Strötgen, Thang Vu, and Dan Zhang for helpful discussions and feedback on
the
writing.
We also thank Jason Wei and Jiaqi Zeng for answering questions about their work.

\section*{Limitations}

This paper is a survey, structuring, organizing and describing works and concepts to 
address
\textit{class imbalance} including \textit{long-tailed learning}.
While we touch upon \textit{data augmentation} and \textit{few-shot learning}, we do not comprehensively review those areas.
Details on the scope of this review have also been described in \sref{sec:intro}.

The search process for the survey included searching for the keywords \textit{class imbalance} and \textit{long tail} in Google Scholar and the ACL Anthology, 
as well as
carefully checking the papers that cite
relevant papers.

Finally, the paper only constitutes a literature review, it does not yet provide a comprehensive 
empirical
study which is much needed in this research area, but it will be of use in carrying out such a study.

\bibliography{anthology,custom}
\bibliographystyle{acl_natbib}

\newpage
\appendix

\section*{Appendix}

\section{Method Overview}
\label{sec:appendix}
We here provide details on a selection of methods surveyed in this paper.
\tref{tab:overview} shows whether they have been applied respectively whether they are applicable in binary, multi-class, and multi-label classification.
Moreover, it contains information on whether authors open-sourced their implementation.
For links to open-sourced code, see \tref{tab:code}.

\section{Practical advice}
\label{sec:practical}

\blue{In \fref{fig:flowchart}, we provide practical advice which class imbalance methods might be beneficial under which circumstances.
Due to the lack of an established benchmark, we can only give rough guidance.}

\begin{table}[h]
\footnotesize
\centering
\begin{tabular}{l|l}
\toprule
     \textbf{Method} & \textbf{Link} \\
     \toprule
     \textit{Data Augmentation} \\
     EDA \citep{wei-zou-2019-eda} & \texttt{\href{https://github.com/jasonwei20/eda_nlp}{GitHub}} \\
     GE3 \citep{wei-2021-good} &
     \texttt{\href{https://aclanthology.org/attachments/2021.emnlp-main.479.Software.zip}{ACL Anthology}} \\
     ECRT \citep{chen2021supercharging} & \texttt{\href{https://github.com/ZidiXiu/ECRT}{GitHub}} \\
     \midrule
     \textit{Loss Functions} \\
     FL \citep{lin2017focal} & \texttt{\href{https://github.com/facebookresearch/fvcore/blob/main/fvcore/nn/focal_loss.py}{GitHub}} \\
     ADL \citep{li-etal-2020-dice} & \texttt{\href{https://github.com/ShannonAI/dice_loss_for_NLP}{GitHub}}\\
     LDAM \citep{cao2019ldam} & \texttt{\href{https://github.com/kaidic/LDAM-DRW}{GitHub}} \\
     DB \citep{wu2020distribution} & \texttt{\href{https://github.com/wutong16/DistributionBalancedLoss}{GitHub}}\\
     \midrule
     \textit{Staged Learning} \\
     cRT \citep{kang2020decoupling} & \texttt{\href{https://github.com/facebookresearch/classifier-balancing}{GitHub}} \\
     ST \citep{jang2021sequential} & \texttt{\href{https://github.com/joeljang/Sequential-Targeting}{GitHub}} \\
     \midrule
     \textit{Model Design} \\
     $\tau$-norm \citep{kang2020decoupling} & \texttt{\href{https://github.com/facebookresearch/classifier-balancing}{GitHub}} \\
     ProtoBERT \citep{tanzer-etal-2022-memorisation} &  \texttt{\href{https://github.com/Michael-Tanzer/BERT-mem-lowres}{GitHub}} \\
     \bottomrule
\end{tabular}
\caption{\textbf{Open-sourced implementations} of examples of class imbalance methods.}
\label{tab:code}
\end{table}

\begin{table*}
\footnotesize
\centering
\setlength\tabcolsep{3pt}
\renewcommand*{\arraystretch}{1.1}
\begin{tabular}{l|c|c|c|c}
\toprule
     Method & Binary classification & Multi-class classification & Multi-label classification & Code \\
     \toprule
     \multicolumn{5}{c}{\textit{Re-Sampling}} \\\midrule
     ROS/RUS (\sref{sec:sampling}) & \checkmark & \checkmark & ? & N/A\\\midrule
     CAS \citep{li2016class-aware} & \checkmark & \checkmark & ? & \texttimes\\
     \toprule
     \multicolumn{5}{c}{\textit{Data Augmentation}} \\\midrule
    EDA \citep{wei-zou-2019-eda} & \citet{juuti-etal-2020-little} & \citet{jiang2021textcut} & \citet{zhang2022attention_text_augmentation, zhang2020data} & \checkmark \\
    & \citet{jiang2021textcut} & & & \\\midrule
    TextCut \citep{jiang2021textcut} & \citet{jiang2021textcut} & \citet{jiang2021textcut} & \checkmark & \texttimes\\\midrule
    GE3 \citep{wei-2021-good} & \checkmark & \citet{wei-2021-good}  & ? & \checkmark\\
     & & \citet{wei2022reprint} & & \\\midrule
     MISO \citep{tian-etal-2021-embedding} & \citet{tian-etal-2021-embedding} & \citet{tian-etal-2021-embedding}* & ? & \texttimes\\\midrule
     ECRT \citep{chen2021supercharging} & \checkmark & \checkmark & \citet{chen2021supercharging}  &  \checkmark \\
     \toprule
     \multicolumn{5}{c}{\textit{Loss Functions}} \\\midrule
     WCE (\sref{sec:loss}) & \citeauthor{tayyar-madabushi-etal-2019-cost} & \citet{adel-etal-2017-ranking} & N/A & N/A\\
     & (\citeyear{tayyar-madabushi-etal-2019-cost}) & \citet{li-xiao-2020-syrapropa} & \\\midrule
     FL \citep{lin2017focal} & \checkmark & \citet{li-etal-2020-dice} & \checkmark & \checkmark\\
     & & \citet{nan-etal-2021-uncovering} & & \\\midrule
     ADL \citep{li-etal-2020-dice} & \checkmark & \citet{li-etal-2020-dice} & \checkmark & \checkmark\\
     & & \citet{spangher-etal-2021-multitask} & \\\midrule
     LDAM \citep{cao2019ldam} & \citet{cao2019ldam} & \checkmark & \citet{biswas2021ldam_multilabel} & \checkmark\\
     & \citet{subramanian-etal-2021-fairness} & & & \\\midrule
     WBCE (\sref{sec:loss}) & \checkmark & \texttimes & \citet{yang-etal-2020-hscnn} & N/A \\\midrule
     RL \citep{dos-santos-etal-2015-classifying} & \checkmark & \citet{adel-etal-2017-ranking} & \texttimes & \texttimes \\\midrule
     DB \citep{wu2020distribution} & \texttimes & \texttimes & \citet{huang-etal-2021-balancing} & \checkmark\\
     \toprule
     \multicolumn{5}{c}{\textit{Staged Learning}} \\\midrule
     cRT \citep{kang2020decoupling} & \checkmark & \citet{nan-etal-2021-uncovering} & \checkmark & \checkmark \\
     & & \citet{hu2022event} & & \\\midrule
     ST \citep{jang2021sequential} & \citet{jang2021sequential} & \citet{jang2021sequential} & \checkmark & \checkmark \\
     \toprule
     \multicolumn{5}{c}{\textit{Model Design}} \\
     \midrule
     $\tau$-norm \citep{kang2020decoupling} & \checkmark & \citet{nan-etal-2021-uncovering} & \checkmark & \checkmark  \\\midrule
     SetConv \citep{gao-etal-2020-setconv} & \citet{gao-etal-2020-setconv} &  \citet{gao-etal-2020-setconv} & \checkmark & \texttimes \\ \midrule
     ProtoBERT \citep{tanzer-etal-2022-memorisation} & \checkmark & \citet{tanzer-etal-2022-memorisation} & \checkmark & \checkmark \\\midrule
     HSCNN \citep{yang-etal-2020-hscnn} & \checkmark & \checkmark & \citet{yang-etal-2020-hscnn}  & \texttimes \\
     \bottomrule
\end{tabular}
     \caption{\textbf{Examples of class imbalance methods and NLP application settings.}
     \checkmark: \textbf{method} applicable (but no particular reference reporting experimental results exists)/\textbf{code}: authors open-sourced their implementation, ?: application not straightforward / open research issues.
     *: The authors select only one class as the minority class in their experiments.
     For links to open-sourced code, see \tref{tab:code}.
          Methods for binary and multi-class classification are in general applicable to classification-based \textbf{relation extraction} approaches; applying class-imbalance techniques to \textbf{sequence labeling} methods in general is similar to the case of multi-label classification.
     For example, if sampling for a particular category, the whole sequence sample may contain additional annotations for other categories.
     }
    
     \label{tab:overview}
\end{table*}

\tikzstyle{decision} = [diamond, draw, fill=blue!20,
    text width=4.5em, text badly centered, node distance=2.5cm, inner sep=0pt]
\tikzstyle{block} = [rectangle, draw, fill=blue!10,
    text width=5em, text centered, rounded corners, minimum height=4em]
\tikzstyle{line} = [draw, very thick, color=black!50, -latex']
\begin{figure*}[h]
\centering%
\footnotesize%
    \begin{tikzpicture}[node distance = 2cm, auto]
        \node [decision] (classification_type) {classifica-tion type?};
        \node [decision, right of=classification_type, node distance=4cm] (multilabel_importance) {all classes equally important?};
        \node [block, right of=multilabel_importance, node distance=3cm] (db) {distribution-balanced loss};
        \node [block, below right of=multilabel_importance, node distance=3cm] (wbce) {weighted (binary) CE loss upweighting only classes of interest};
        \node [decision, below of=classification_type, node distance=3cm] (resources) {limited computational resources?};
        \node [decision, below right of=resources] (importance) {all classes equally important?};
        \node [decision, below of=importance, node distance=3cm] (catch_all) {catch-all class?};
        \node [block, below right of=catch_all, node distance=2.5cm] (ranking) {ranking loss};
        \node [block, right of=catch_all, node distance=2.5cm] (wce) {weighted CE loss};
        \node [decision, below left of=resources] (capacity) {limited implementation capacities?};
        \node [block, left of=capacity, node distance=3cm] (ros) {random oversampling};
        \node [decision, below of=capacity, node distance=3cm] (representations) {access to good representations?};
        \node [block, below left of=representations, node distance=2.5cm] (ge3) {GE3};
        \node [block, below right of=representations, node distance=2.5cm] (eda) {EDA};

        \path [line] (classification_type) -- node [, color=black] {multi-label} (multilabel_importance);
        \path [line] (multilabel_importance) -- node [, color=black] {yes} (db);
        \path [line] (multilabel_importance) -- node [, color=black] {no} (wbce);
        
        \path [line] (classification_type) -- node [, color=black] {single-label} (resources);
        \path [line] (resources) -- node [, color=black] {yes} (importance);
        \path [line] (importance) -- node [, color=black] {no} (wbce);
        \path [line] (importance) -- node [, color=black] {yes} (catch_all);
        \path [line] (catch_all) -- node [, color=black] {yes} (ranking);
        \path [line] (catch_all) -- node [, color=black] {no} (wce);
        
        \path [line] (resources) -- node [, color=black] {no} (capacity);
        \path [line] (capacity) -- node [, color=black] {yes} (ros);
        \path [line] (capacity) -- node [, color=black] {no} (representations);
        
        \path [line] (representations) -- node [, color=black] {yes} (ge3);
        \path [line] (representations) -- node [, color=black] {no} (eda);

    \end{tikzpicture}
    \caption{\textbf{Practical advice} which methods to try under which circumstances.}
    \label{fig:flowchart}
\end{figure*}
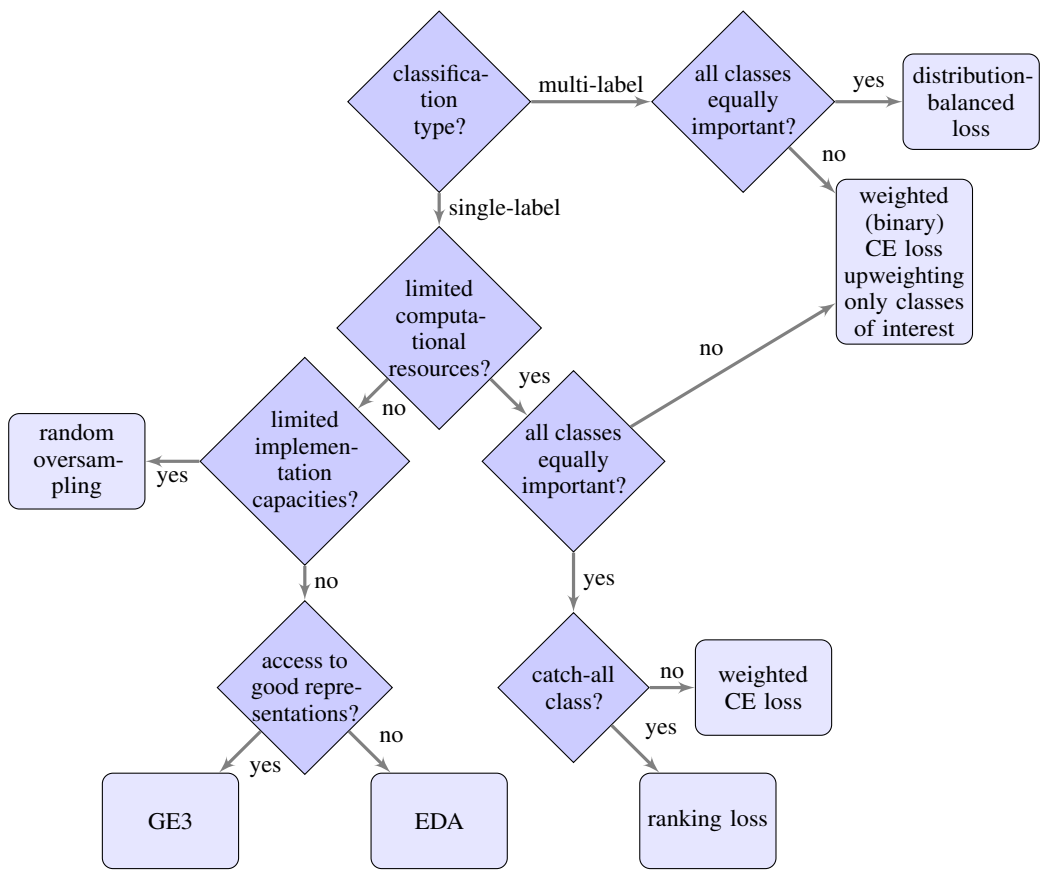

\end{document}